# PatchSTG: Scalable Spatiotemporal Graph Transformers for Traffic Forecasting on Irregular Sensor Networks

Jichao Li[1], Xuanming Shi[2*]

1. Portsmouth Abbey School

2. CodingFuture (Shanghai) Education Technology Co., Ltd.

E-mail: 568078145@qq.com

**Abstract**

Traffic forecasting is a fundamental component of intelligent transportation systems, yet remains challenging in real-world settings due to irregular sensor distributions and the high computational cost of modeling large-scale spatiotemporal dependencies. In practical traffic networks, sensors are unevenly distributed across regions, leading to non-uniform spatial structures that limit the effectiveness and scalability of existing graph-based and attention-based models.To address these challenges, we propose **PatchSTG**, a patch-based spatiotemporal graph Transformer designed for efficient forecasting on irregular sensor networks. The key idea is to introduce a hierarchical spatial representation that partitions sensors into balanced, locality-preserving patches based on geographic information. On top of this structure, a dual attention encoder alternates between intra-patch attention for capturing local interactions and inter-patch attention for modeling global dependencies, reducing computational complexity from quadratic to near-linear scaling.We evaluate PatchSTG on real-world traffic data from Rhode Island and additional large-scale datasets. Experimental results demonstrate that the proposed model achieves stable and competitive forecasting performance across multiple horizons, while significantly improving computational efficiency. Ablation studies further validate the effectiveness of spatial partitioning and dual attention in capturing both local and long-range traffic dynamics.These results suggest that patch-based spatiotemporal modeling provides a scalable and effective framework for traffic forecasting under irregular spatial settings.

## 1. Introduction

Accurate traffic forecasting is increasingly essential for modern intelligent transportation systems（ITS）, given its significant societal and economic impact. INRIX's 2024 report estimates that congestion cost the U.S. over $74 billion, with major-city drivers losing 40–100 hours annually [1]. Environmentally, transportation contributes 28–29% of national greenhouse gas emissions—the largest or second-largest share—according to EPA inventories [2][3]. Freight systems are also heavily

affected, as congestion imposed $108.8 billion in losses on trucking in 2022 [4]. These findings highlight traffic forecasting as a critical lever for efficiency and public welfare.

This urgency has driven widespread ITS deployment relying on real-time prediction. Systems like SURTRAC reduce travel times by over 25% through decentralized optimization [5]. In emergency response, EMVLight applies multi-agent RL to cut response times by 42.6% [6], with later work showing further gains in congested cities [7]. Infrastructure-integrated methods (e.g., sensor-triggered preemption) [8] and large-scale coordination systems such as SCATS demonstrate network-wide benefits. Across these applications, accurate and scalable traffic forecasting underpins decision-making in routing, signal control, and emergency management.

Despite these advances, traffic flow modeling remains a challenging **spatiotemporal learning problem over large-scale, irregularly structured graphs**. Classical approaches such as ARIMA and Kalman filtering lack the ability to capture spatial dependencies. Deep learning models—including GCN [9], STGCN [10], RNN-based methods [11], and DCRNN [12]—have significantly improved the modeling of spatiotemporal correlations, while recent Transformer-based architectures [13] further enhance long-range temporal dependency learning.

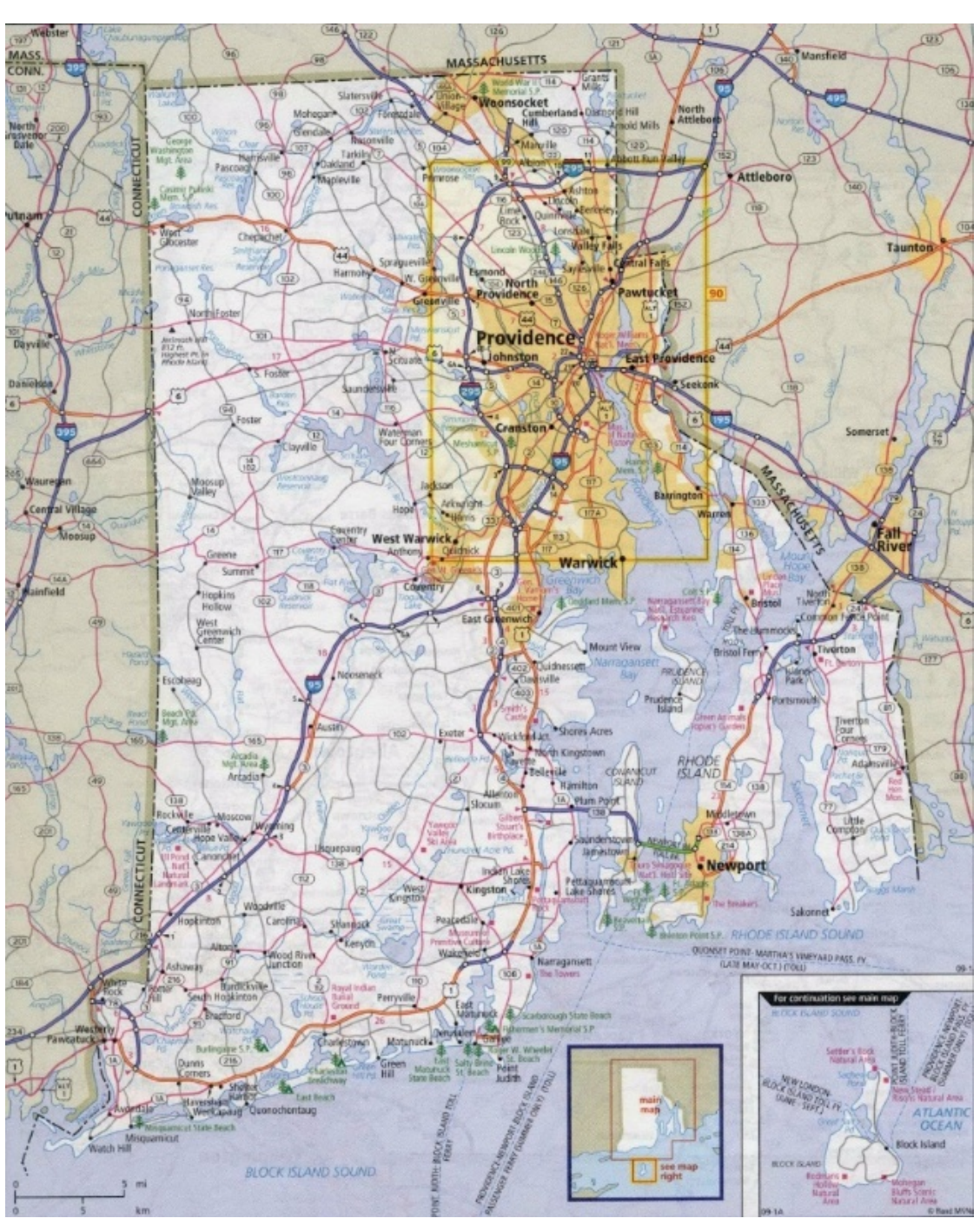

Figure 1: Rhode Island Traffic Map

However, a fundamental challenge arises from **spatial heterogeneity and non-uniform sensor distributions in real-world traffic networks**. As shown in Figure 1, sensor deployments are typically dense around bottlenecks (e.g., freeway interchanges, major bridges) and sparse in suburban or rural regions. Rhode Island exemplifies this pattern: dense coverage along I-95, I-195, and the Claiborne Pell

Newport Bridge stands in contrast to sparse coverage across western and northern municipalities. Such **non-uniformly sampled traffic graphs** present significant challenges for existing models. Grid-based representations often lead to empty or overloaded cells, while fixed adjacency matrices fail to capture dynamically evolving spatial relationships.

In addition, existing deep traffic models suffer from **limited scalability and interpretability**. Many graph-based and Transformer-based approaches incur quadratic complexity with respect to the number of sensors, making them difficult to apply to large-scale networks. Meanwhile, their black-box nature limits their usability in real-world traffic management, where operators must understand whether predicted congestion arises from periodic demand patterns, upstream propagation, or structural bottlenecks.

To address these challenges, we propose **PatchSTG**, a patch-based spatiotemporal graph Transformer designed for large, irregular sensor networks. The key idea is to introduce a **hierarchical patch-based representation** that decomposes the original traffic graph into balanced, locality-preserving substructures. Specifically, PatchSTG employs an improved Leaf KDTree to partition sensors into spatially coherent patches, enabling efficient localized modeling. A Dual Attention Encoder alternates between Depth Attention within patches and Breadth Attention across patches, reducing complexity from $O(N^2)$ to $O(N \cdot P + N^2/P)$, achieving $O(N^{1.5})$ when $P \approx \sqrt{N}$. Moreover,the hierarchical patch structure also provides natural interpretability by exposing local congestion propagation and long-range cross-patch interactions.

We evaluate PatchSTG on Rhode Island, a representative mid-scale traffic network characterized by complex geography, critical infrastructure bottlenecks, and highly irregular sensor distributions. supported by high-quality GIS layers and traffic datasets from RIDOT [18]. These characteristics make it a compelling real-world testbed for assessing scalability and robustness. In addition, we conduct extensive experiments on three additional large-scale traffic datasets to validate generalizability

Our contributions are summarized as follows:

1) We formulate traffic forecasting over irregular sensor networks as a problem of learning on **non-uniform spatiotemporal graphs**, highlighting the challenges of spatial heterogeneity and scalability.
2) We propose a **patch-based spatiotemporal graph Transformer** that reduces computational complexity from $O(N^2)$ to $O(N^{1.5})$ while preserving modeling capacity.
3) We introduce a hierarchical attention mechanism that provides **interpretable insights** into both local and global traffic dynamics.
4) We demonstrate through extensive experiments that PatchSTG achieves strong predictive performance while maintaining efficiency and robustness across diverse datasets..

## 2. Literature Review

### 2.1 Surveys on Traffic Prediction, Wireless Sensor Networks, and Large-Scale Streaming Analytics

Survey studies outline systemic challenges in large-scale traffic prediction. Yuan and Li [19] identify persistent gaps—limited real-time scalability, high computational cost, and difficulty modeling heterogeneous spatial structures. In parallel, WSN-focused surveys highlight infrastructure limitations: Nellore and Hancke [20] show that although WSNs support local sensing, scalable city-wide prediction remains difficult due to resource constraints and irregular sensor deployments.

From a data systems perspective, Djedouboum et al. [21] further emphasize that large-scale sensor streams strain low-latency data collection and aggregation frameworks, especially under heterogeneous node density. Similarly, Medeiros et al. [22] demonstrate that spatial non-uniformity and non-stationary data distributions lead to model drift in online learning systems, degrading predictive performance over time. **Collectively, these surveys point to a shared conclusion:** large-scale traffic forecasting is constrained by three fundamental challenges—limited real-time scalability, high computational cost, and the difficulty of modeling irregular, non-uniform sensor networks. These issues define the core problem setting addressed in this work.

### 2.2 Deep Spatiotemporal Models for Traffic Forecasting

Deep spatiotemporal models have significantly advanced traffic forecasting by capturing complex temporal dynamics and spatial dependencies.Graph-based deep learning models significantly advance traffic forecasting but remain limited in scalability. STGCN [23], DCRNN [24], and Graph WaveNet [25] improve spatiotemporal modeling through graph convolutions and diffusion mechanisms, while attention-based architectures—ASTGCN [26], STTN [27], and TFT [28]—capture dynamic spatial–temporal correlations with enhanced interpretability.

More recent work explores large-scale attention mechanisms and global modeling paradigms. For instance,TimeSformer [29] and global-scale models like GraphCast [30] show the potential of large learned models for high-dimensional forecasting. GMAN [31] and ST-Transformer [32] further integrate multi-attention mechanisms and adaptive adjacency learning to capture complex relational structures.

**Despite these advances, existing methods share several critical limitations.** First, many attention-based models incur quadratic computational complexity with respect to the number of sensors, limiting their scalability to large networks. Second, modeling global spatial dependencies remains computationally expensive and often redundant. Third, and most importantly, **irregular spatial layouts and non-uniform sensor distributions are not explicitly addressed**, as most methods rely on fixed graph structures or uniform assumptions. These limitations motivate the need for scalable and structure-aware architectures that can operate effectively on non-uniform spatiotemporal graphs.

### 2.3 Spatial Partitioning, KD-Tree, and Patch-Based Modeling

To address irregular spatial structures, prior work has explored spatial partitioning and hierarchical representations. Wei et al. [33] use N-KD Trees to generate balanced, proximity-preserving spatial groups, while Wang [34] enhances KD-tree storage with density estimation and locality-sensitive hashing（LSH） to enable efficient querying over large, uneven spatial datasets. These approaches demonstrate the effectiveness of hierarchical spatial decomposition in managing non-uniform data distributions.

At the graph representation level, PatchGT [35] introduces a patch-based Transformer pipeline operating on clustered graph partitions, demonstrating that patch-level modeling can improve scalability without sacrificing global relational learning. Such approaches suggest that **hierarchical partitioning combined with localized modeling** is a promising direction for scaling graph-based learning.

**However, existing partitioning and patch-based methods are not specifically designed for spatiotemporal traffic forecasting**, where both dynamic temporal dependencies and irregular spatial structures must be jointly modeled. This gap motivates the development of a unified framework that integrates spatial partitioning with spatiotemporal learning.

Taken together, existing traffic forecasting methods struggle to jointly achieve scalability, computational efficiency, and robustness to irregular spatial structures. These limitations are fundamentally coupled, as high computational cost restricts large-scale deployment while uniform structural assumptions hinder performance on real-world, non-uniform sensor networks. This gap motivates the need for structure-aware and scalable spatiotemporal models. Our proposed patch-based spatiotemporal graph Transformer addresses this challenge by integrating hierarchical spatial partitioning with efficient attention mechanisms, enabling accurate and scalable forecasting on irregular sensor graphs.

## 3. Problem Definition

### 3.1 Traffic Data

Traffic data consist of multiple correlated time series collected from road-side sensors deployed across an urban traffic network. Each sensor is associated with a fixed spatial location and continuously records traffic-related measurements, such as traffic flow or speed, over time.

Formally, the recorded traffic signal at sensor $n$ is represented as a univariate time series $x_n \in R^H$, where $H$ denotes the number of historical time steps. Aggregating observations from all $N$ sensors, the traffic dataset can be organized as a matrix

$$\boldsymbol{X} \in \boldsymbol{R}^{\boldsymbol{H} \times \boldsymbol{N}}$$

where $x_h^n$ represents the traffic measurement observed at sensor $n$ during time step $h$.

In addition to temporal observations, each sensor is associated with real-world geographic coordinates. Let

$$\mathbf{Lat} \in \mathbb{R}^{\boldsymbol{N}}, \mathbf{Lng} \in \mathbb{R}^{\boldsymbol{N}}$$

denote the latitude and longitude of all sensors, respectively. These spatial attributes provide essential information for modeling spatial dependencies in traffic dynamics.

## 3.2 Traffic Forecasting Task

The objective of traffic forecasting is to predict future traffic conditions based on historical observations and spatial information. Specifically, given traffic measurements from the preceding $H$ time steps, the task is to forecast traffic values for the next $F$ time steps across all sensors.

Formally, the forecasting process can be defined as

$$\hat{\mathbf{Y}} = f_\theta(\mathbf{X}, \mathbf{Lat}, \mathbf{Lng}),$$

Where $\hat{\mathbf{Y}} \in \mathbb{R}^{F \times N}$ denotes the predicted future traffic, $f_\theta(\cdot)$ represents a data-driven forecasting model parameterized by $\theta$, and the predictions are evaluated against the ground-truth future observations

$$\mathbf{Y} \in \mathbb{R}^{F \times N}.$$

This formulation captures the joint temporal and spatial nature of traffic forecasting and serves as a unified problem definition for a wide range of spatiotemporal modeling approaches.

## 3.3 Problem Characteristics

Although the above formulation appears straightforward, real-world traffic forecasting poses several fundamental challenges.

**Irregular Spatial Distribution:** Traffic sensors are irregularly distributed over road networks, with dense deployments around highways, interchanges, and bottlenecks, and sparse coverage in suburban or rural regions. Such spatial heterogeneity makes regular grid-based representations unsuitable and complicates the modeling of spatial dependencies.

**Scalability:** Modern urban traffic networks may contain thousands of sensors, leading to high computational and memory demands for models that explicitly capture pairwise spatial interactions. Designing forecasting models that scale efficiently with network size remains a critical challenge.

**Interpretability:** For practical deployment in intelligent transportation systems, forecasting models are expected not only to achieve high predictive accuracy but also to provide interpretable spatial dependencies. Traffic operators must be able to understand how congestion propagates across different regions of the network and which spatial interactions contribute most to future predictions.

These challenges motivate the development of scalable, structure-aware, and interpretable spatiotemporal forecasting models, which we address in the following sections.

# 4. Methodology

## 4.1 Overall Architecture

We propose a patch-based spatiotemporal forecasting framework that models traffic dynamics through hierarchical spatial decomposition and dual-level attention mechanisms. The overall architecture of the proposed model is illustrated in Figure 2, which follows a hierarchical and modular design to efficiently model large-scale traffic networks. Given historical traffic observations and sensor locations, the framework consists of four main components: (1) spatiotemporal embedding, (2) irregular spatial partitioning, (3) a dual attention encoder, and (4) a projection decoder. These components are organized in a coarse-to-fine manner, progressively transforming raw traffic signals into structured representations that capture both local and global spatiotemporal dependencies.

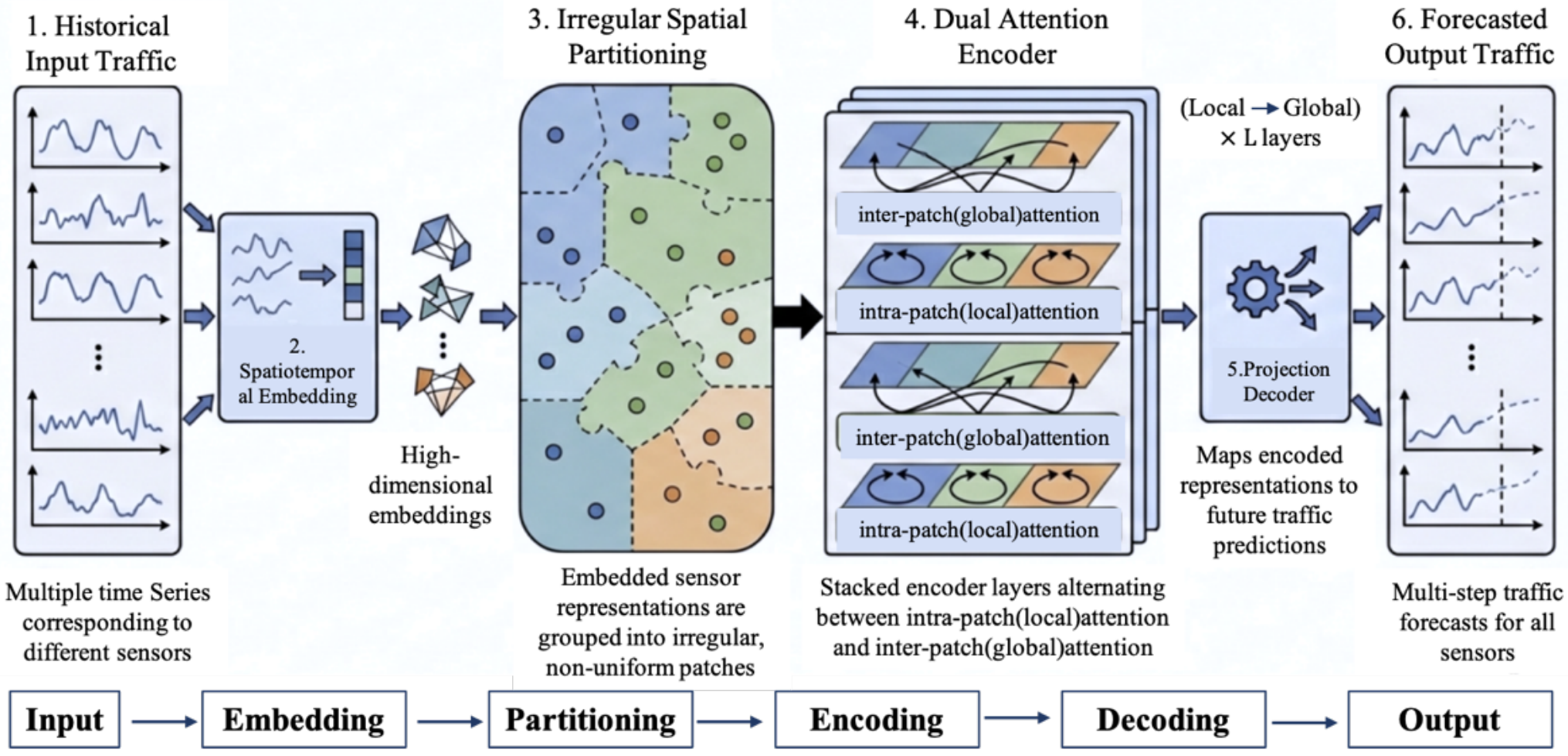

Figure 2: Overall architecture of the proposed spatiotemporal forecasting framework. The model takes historical traffic observations as input and first applies a spatiotemporal embedding module to obtain high-dimensional representations. These embeddings are then partitioned into irregular spatial patches based on sensor locations. A dual attention encoder alternates between intra-patch (local) attention and inter-patch (global) attention to capture spatial dependencies efficiently. Finally, a projection decoder outputs multi-step traffic forecasts for all sensors.

The framework first maps raw traffic time series into a high-dimensional latent space. Specifically, historical traffic observations are embedded to capture temporal patterns and to enhance the distinguishability of individual sensors. This embedding stage serves as a preprocessing step that converts numerical traffic measurements into representations suitable for subsequent spatial modeling.

Instead of modeling spatial dependencies directly on the entire sensor set, the embedded traffic representations are partitioned into a collection of spatial patches. This partition is performed based on sensor locations and is designed to handle irregular sensor distributions commonly observed in real-world traffic networks. By grouping spatially proximate sensors into balanced, non-overlapping patches, the framework reduces the number of elements involved in spatial interaction modeling. This design significantly improves computational efficiency while preserving essential local spatial structure.

On top of the patched representations, a dual attention encoder is employed to capture spatial dependencies at multiple levels. The encoder alternates between two complementary attention mechanisms. First, intra-patch attention is applied within each patch to model fine-grained local spatial interactions among nearby sensors. Then, inter-patch attention is conducted across patches to capture global spatial dependencies and long-range traffic propagation patterns. By decoupling local and global modeling, the encoder effectively balances expressive power and computational cost. Multiple layers of the dual attention encoder can be stacked to progressively refine spatiotemporal representations.

Finally, the learned representations are transformed back to the original sensor space and projected into the future. A projection decoder maps the encoded features to forecast traffic values for the next $F$ time steps at each sensor location. The model is trained end-to-end by minimizing the discrepancy between predicted and ground-truth traffic observations.

### 4.2 Irregular Spatial Partitioning

Urban traffic sensor networks exhibit highly irregular spatial distributions, with dense deployments around highways and bottlenecks and sparse coverage in peripheral regions. Such heterogeneity renders regular grid-based partitioning ineffective and leads to high computational cost when modeling all sensors globally. To address this challenge, we adopt an irregular spatial partitioning strategy that groups sensors into spatially coherent and balanced patches based solely on their geographic locations. This design introduces an intermediate representation between individual sensors and global spatial modeling, enabling efficient and structure-aware learning. This partitioning can be viewed as constructing a hierarchical approximation of the original sensor graph, where local neighborhoods are explicitly preserved while reducing global interaction complexity.

Given embedded sensor representations, we partition the sensor set into $K$ non-overlapping spatial patches

$$\mathcal{P} = \{P_1, P_2, \dots, P_K\},$$

where each patch contains geographically proximate sensors and the union of all patches covers the entire sensor network.

The partitioning is designed to satisfy three properties: spatial coherence, ensuring that local correlations are preserved within each patch; balance, preventing extreme variations in patch sizes; and scalability, reducing the complexity of subsequent global modeling by operating at the patch level rather than the sensor level.

By introducing this patch-based spatial representation, the framework naturally decouples local and global spatial dependencies, providing the structural foundation for the dual attention encoder described in the next section.

### 4.3 Dual Attention Encoder

Building on the patch-based spatial representation, we introduce a dual attention encoder that explicitly models spatial dependencies at two complementary levels: local interactions within patches and global interactions across patches**.** This design decouples fine-grained and coarse-grained spatial modeling, enabling both expressive representation learning and scalable computation.

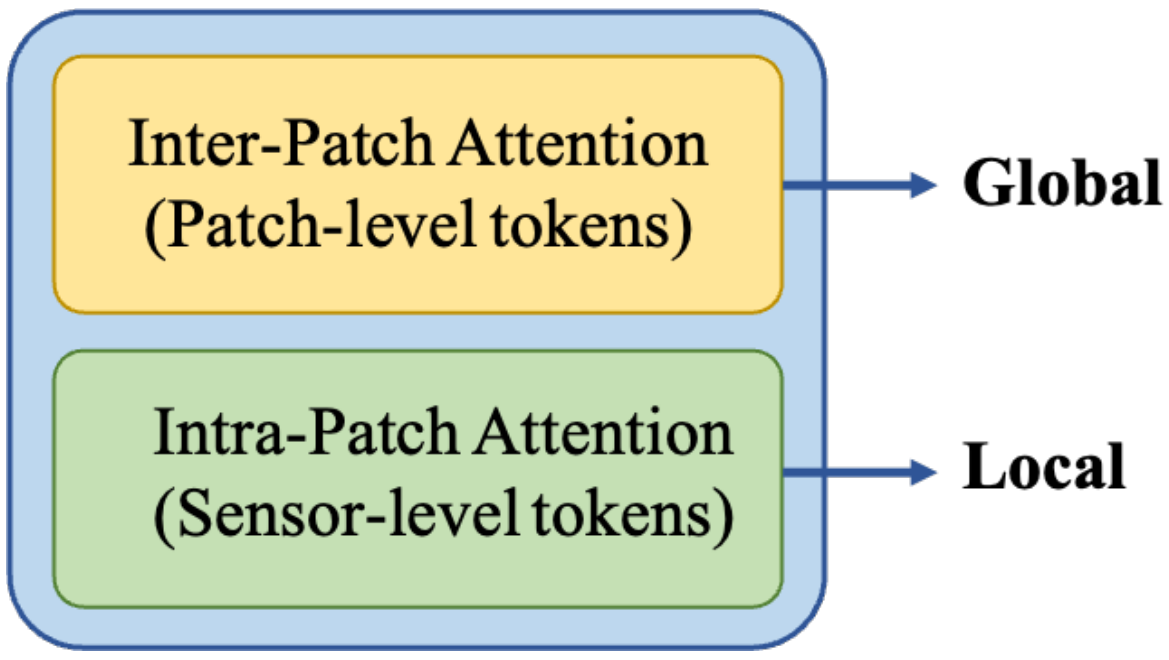

Figure 3: Illustration of the dual attention encoder.

Each encoder layer alternates between intra-patch (local) attention, which models fine-grained spatial interactions within patches, and inter-patch (global) attention, which captures long-range dependencies across patches.

Within each spatial patch, sensors are treated as a local group, and their representations are updated via intra-patch attention. This mechanism captures short-range spatial correlations among geographically proximate sensors, such as congestion propagation along adjacent road segments or within localized regions. By restricting attention computation to sensors within the same patch, intra-patch attention preserves local spatial structure while maintaining a manageable computational cost.

To model long-range spatial dependencies, the encoder further applies inter-patch attention across patches. At this stage, each patch is represented by aggregated embedding, and attention is computed at the patch level rather than the sensor level. Inter-patch attention captures global traffic interactions among distant regions, such as network-wide congestion patterns or coordinated demand fluctuations, while significantly reducing the complexity of global spatial modeling.

The dual attention encoder consists of multiple stacked layers that alternate between intra-patch (local) attention and inter-patch (global) attention. This alternating structure allows information to propagate hierarchically: local patterns are first refined within patches and then exchanged globally across patches. By iteratively integrating local and global spatial context, the encoder produces rich spatiotemporal representations that balance modeling capacity and scalability.Formally, the encoder decomposes spatial modeling into two levels: sensor-level interactions within patches and patch-level interactions across regions, enabling a hierarchical representation of spatial dependencies.

### 4.4 Computational Complexity Analysis

We analyze the computational complexity of the proposed framework with a focus on spatial modeling, which is the dominant cost in large-scale traffic forecasting.

Let $N$ denote the number of sensors, $K$ the number of spatial patches, and $S$ the average number of sensors within each patch, where $N \approx K \times S$. Let $d$ denote the feature dimension.

In conventional dynamic spatial modeling approaches, spatial dependencies are computed directly at the sensor level. Such methods typically involve pairwise interactions among all sensors, resulting in a computational complexity of

$$O(N^2 d),$$

which becomes prohibitive as the size of the traffic network increases.

With irregular spatial partitioning, spatial modeling is decomposed into local and global stages.

Local attention is applied independently within each patch. The total cost across all patches is

$$\mathcal{O}(K \cdot S^2 d).$$

Global attention operates at the patch level, where interactions are computed among patch representations. This results in a complexity of

$$\mathcal{O}(K^2 d).$$

Combining both stages, the overall spatial modeling cost of a single encoder layer is

$$\mathcal{O}(KS^2 d + K^2 d).$$

In practice, the patch size $S$ is bounded and significantly smaller than $N$, while the number of patches $K$ grows sublinearly with respect to $N$. As a result, the dominant cost scales approximately linearly with the number of sensors:

$$\mathcal{O}(NSd),$$

which is substantially more efficient than quadratic sensor-level modeling.

This complexity reduction enables the proposed framework to scale to large urban traffic networks while preserving both local fidelity and global spatial awareness. Thus, the proposed framework reduces the quadratic complexity of global attention to near-linear scaling, enabling efficient modeling of large-scale traffic networks.

## 5. Experiments

In this section, we present experimental results to analyze the characteristics of the traffic data, evaluate the training process of the proposed model, and investigate the effectiveness of different model components through ablation studies.

### 5.1 Data Analysis and Characteristics

Before training the forecasting model, we first conduct exploration data analysis to understand the temporal and spatial characteristics of the traffic dataset. The experiments are based on traffic data collected from the Rhode Island sensor network, which is further divided into four regional subsets to reflect different spatial scales and traffic conditions. Specifically, SD, GBA, GLA, and CA correspond to four representative subregions within Rhode Island, differing in sensor density, spatial coverage, and dominant traffic patterns. Table 1 summarizes the characteristics of these regional subsets used in this project.

Table 1: Description of Regional Subsets in the Rhode Island Traffic Dataset

| Name | Description |
|---|---|
| SD | A suburban-dominant subregion in Rhode Island |
| GBA | A large-scale road network area in Rhode Island, mainly covering major roads |
| GLA | A region in Rhode Island with highly uneven sensor distribution |
| CA | A dense urban core area in Rhode Island |

Figure 4 illustrates the average daily traffic pattern of a representative sensor (Sensor 0). The traffic flow exhibits clear periodic behavior within a day, with pronounced peak and off-peak periods, indicating strong temporal regularity in traffic dynamics.

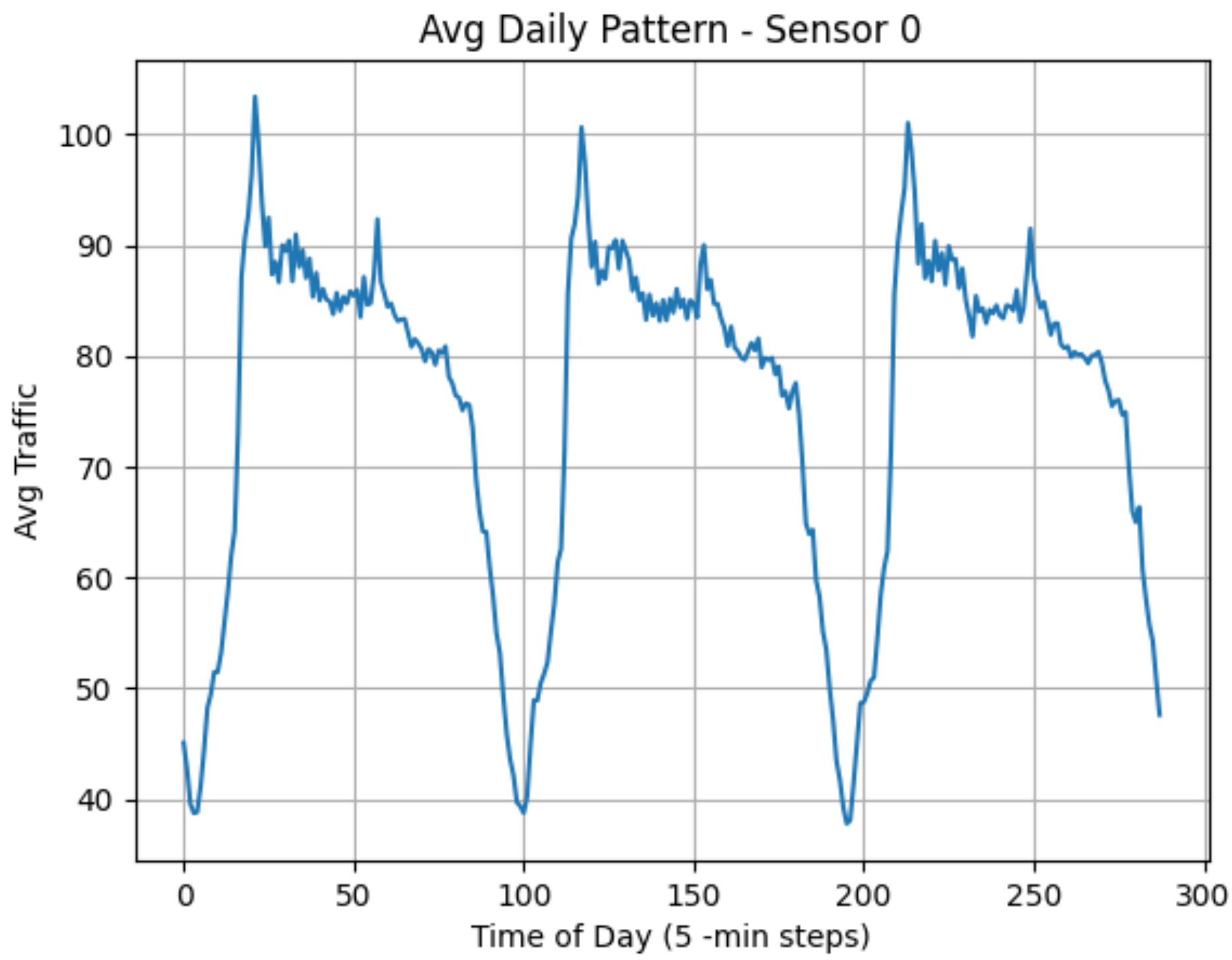

Figure 4: Average daily traffic pattern of Sensor 0 over a 24-hour period (5-minute intervals).

To further examine weekly variations, Figure 5 shows the average daily traffic patterns across different weekdays for the same sensor. While the overall daily trends are consistent, noticeable variations exist among weekdays, suggesting that traffic patterns are influenced by weekly routines.

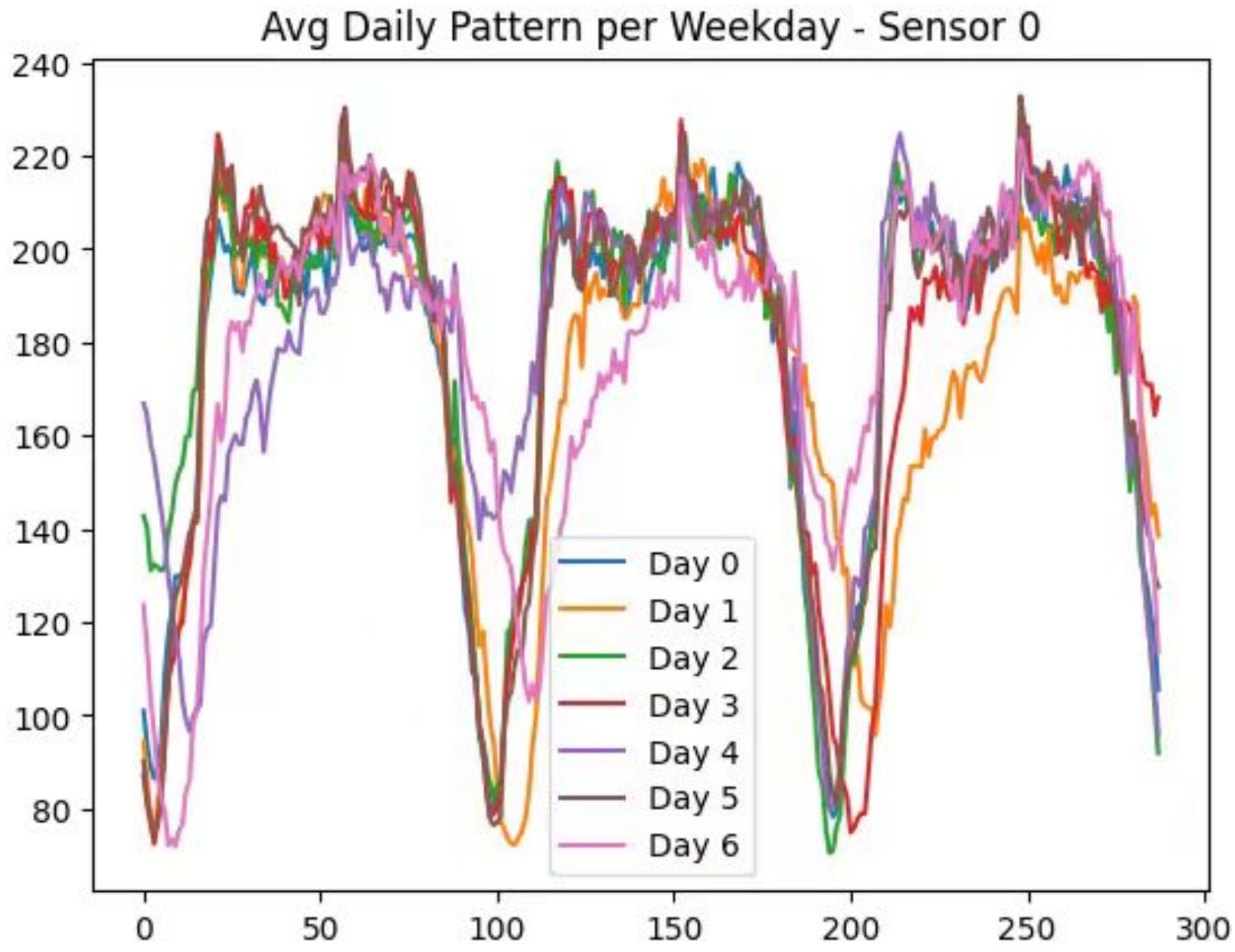

Figure 5: Average daily traffic patterns of Sensor 0 across different weekdays.

Figure 6 presents the distribution of traffic values for Sensor 0. The distribution is multi-modal and skewed, reflecting the coexistence of low-traffic and high-traffic states during different time periods.

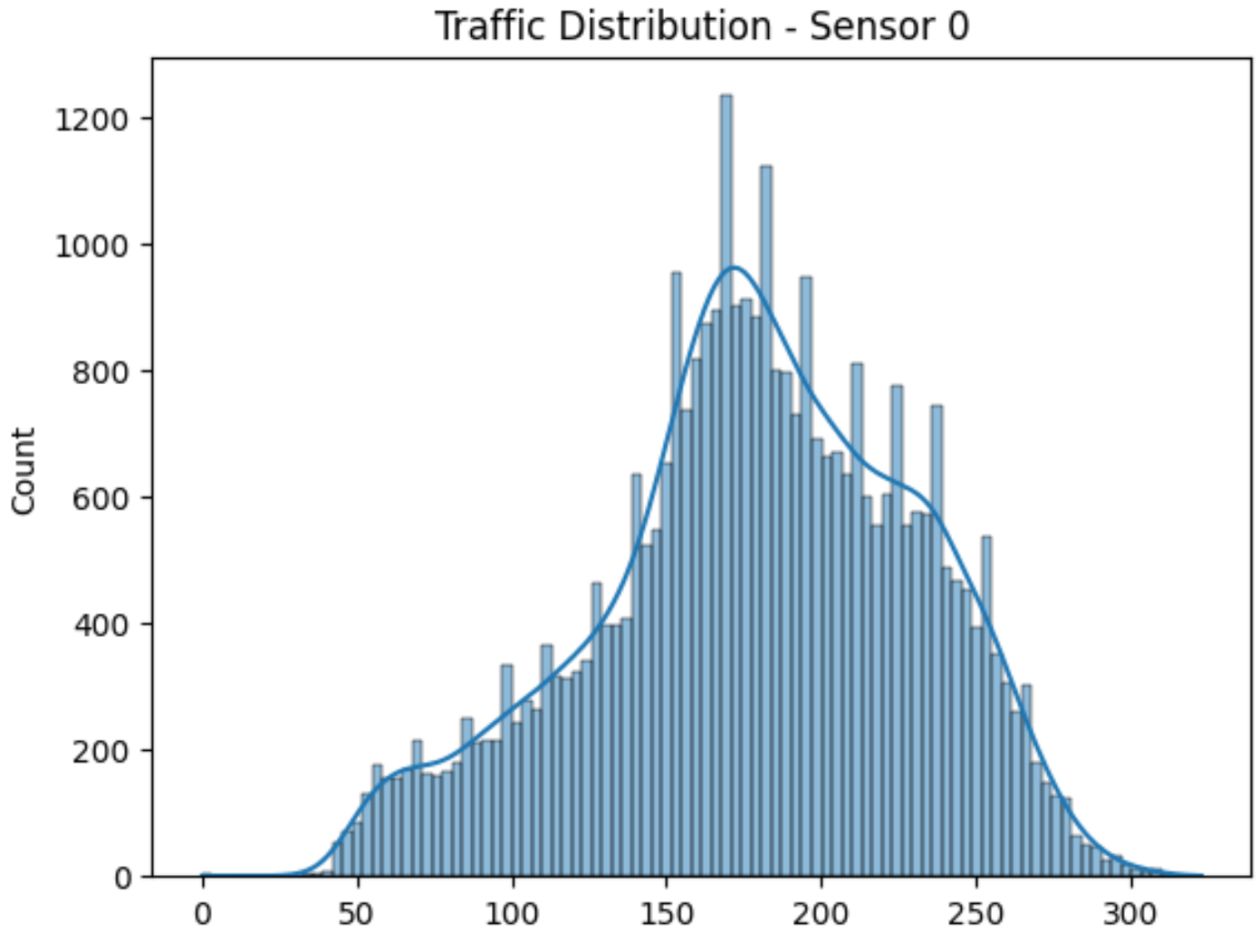

Figure 6: Distribution of traffic values for Sensor 0.

Figure 7 displays traffic time series from multiple sensors over the entire observation period. Although all sensors exhibit similar periodic trends, their traffic magnitudes and fluctuations differ significantly, highlighting spatial heterogeneity across the sensor network.

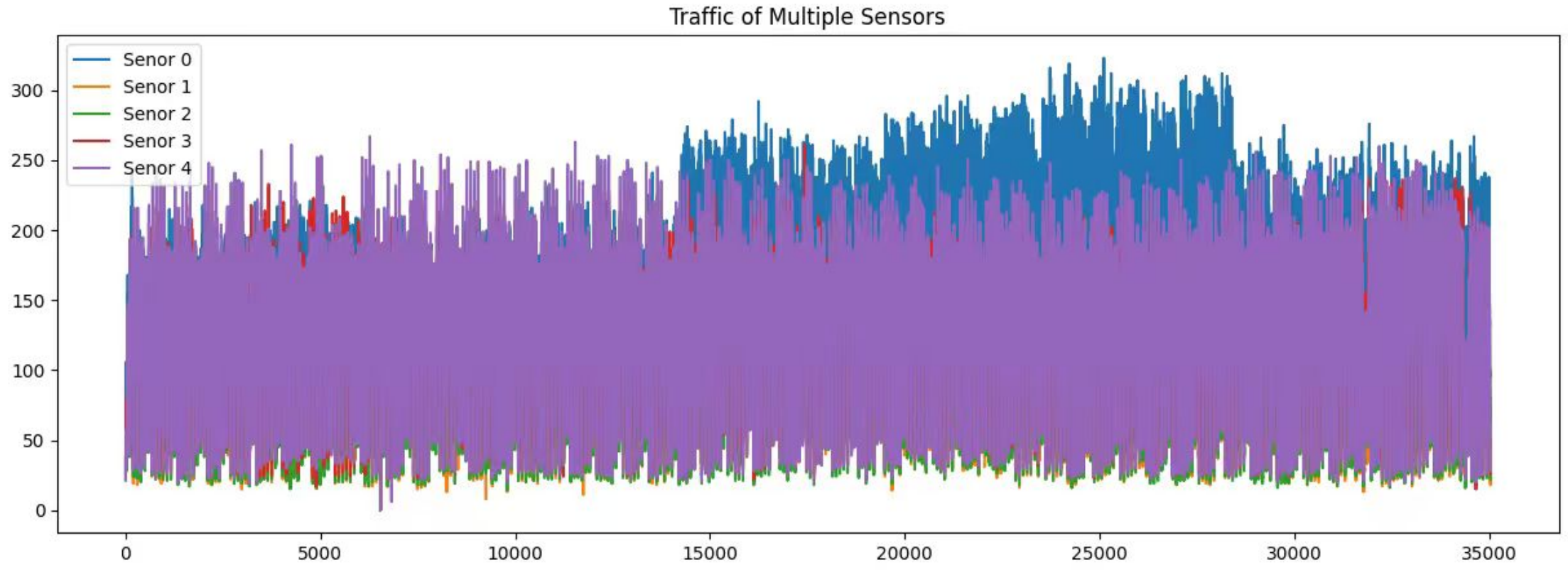

Figure 7: Traffic time series of multiple sensors over time.

Figure 8 visualizes the spatial distribution of traffic sensors, where each point corresponds to a sensor location and the color indicates the average traffic volume. Sensors are unevenly distributed across space, and traffic intensity varies substantially among different regions. This irregular spatial structure motivates the use of spatially adaptive modeling strategies.

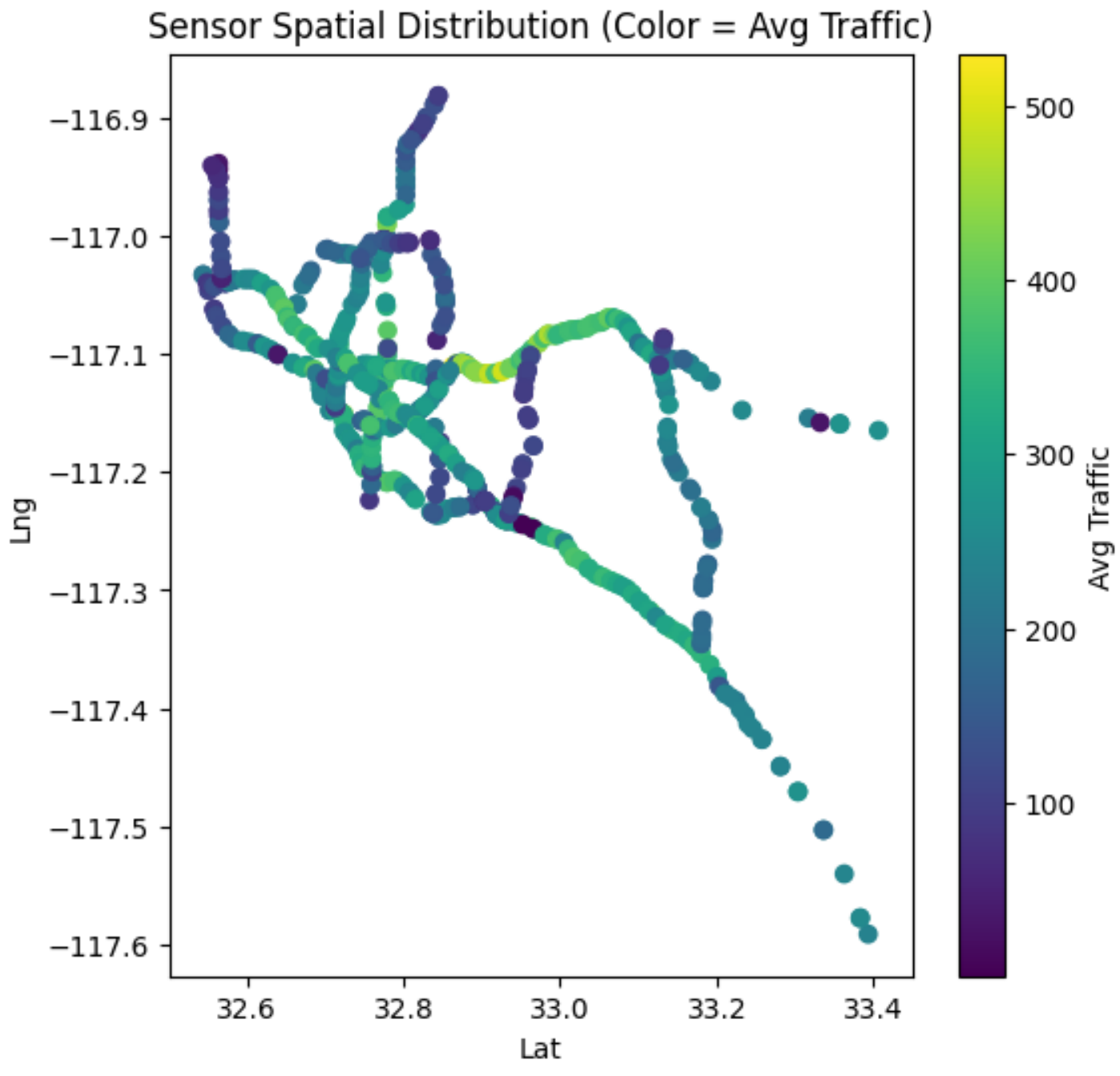

Figure 8: Spatial distribution of sensors colored by average traffic volume.

To further analyze spatial relationships, Figure 9 presents the sensor-to-sensor correlation heatmap. Strong correlations are observed among nearby sensors, indicating significant spatial dependencies in traffic dynamics.

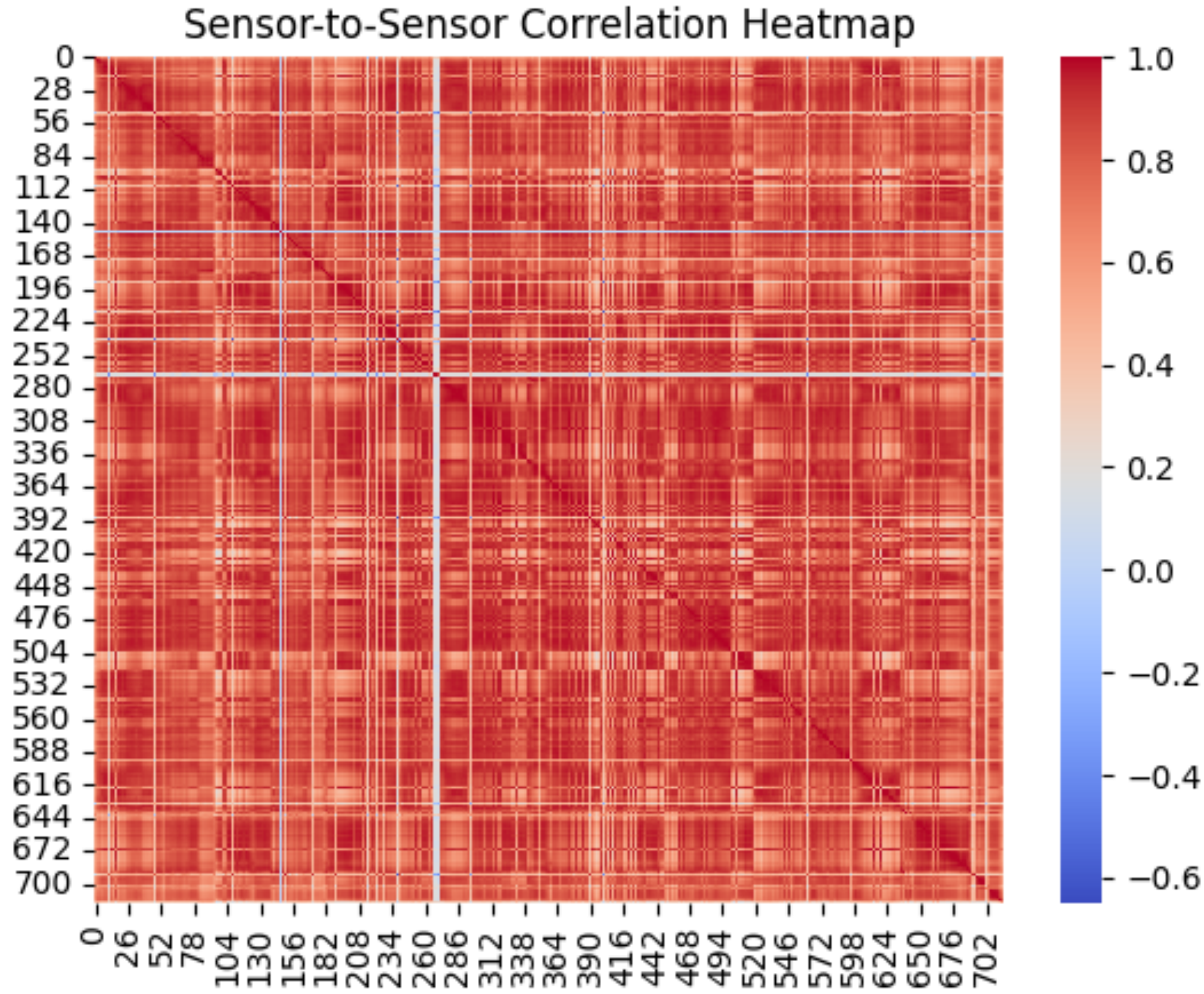

Figure 9: Sensor-to-sensor correlation heatmap.

5.2 Model Training and Optimization

We evaluate the training behavior of the proposed PatchSTG model to ensure stable optimization.

Figure 10 shows the training loss as a function of training epochs. The loss decreases steadily and converges after approximately 40 epochs, demonstrating that the model can be trained efficiently without optimization instability.

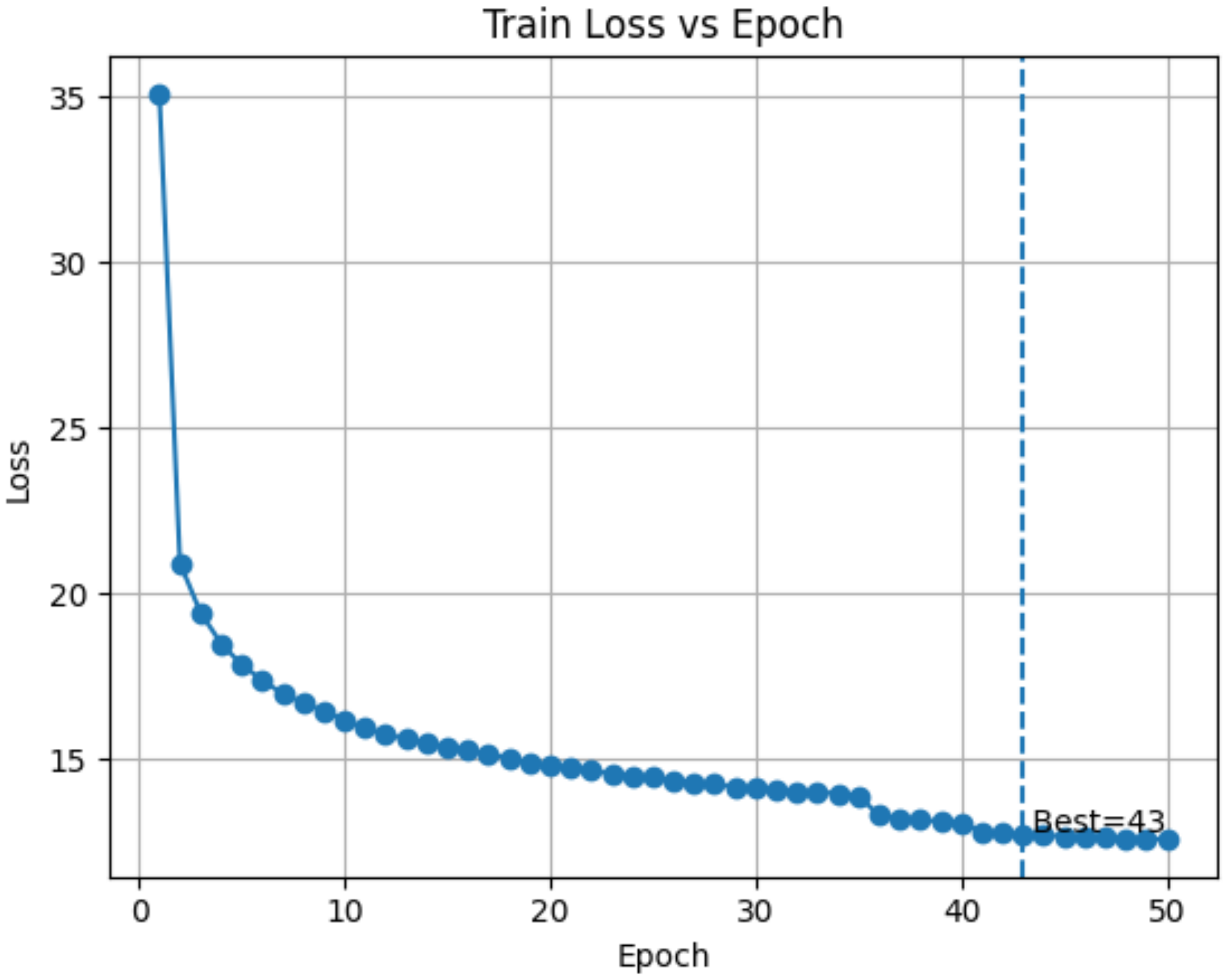

Figure 10: Training loss versus training epoch.

Figure 11 reports the average training metrics, including MAE, RMSE, and MAPE, across epochs. All metrics show consistent improvement during training and gradually stabilize, further confirming the convergence of the learning process.

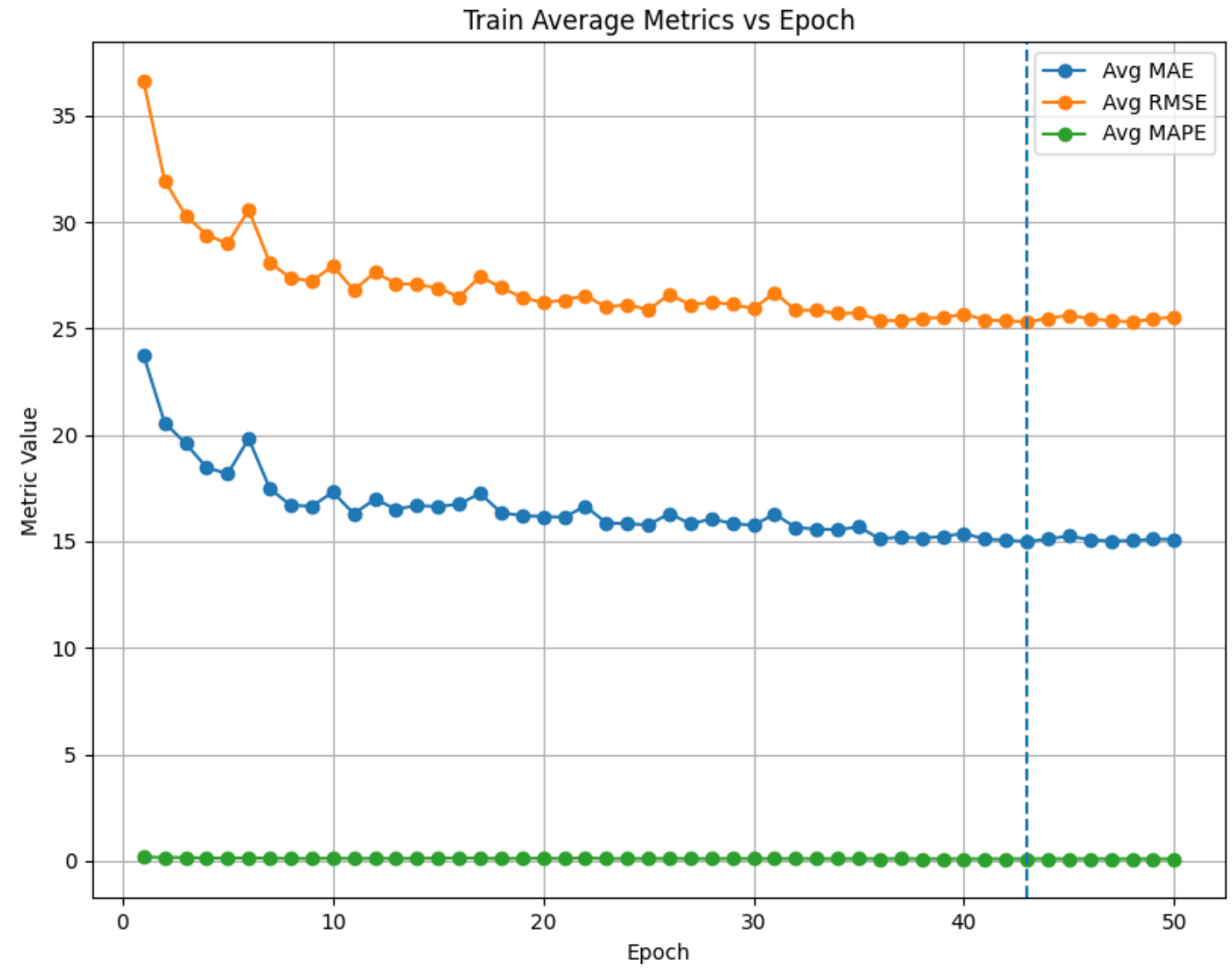

Figure 11: Training MAE, RMSE, and MAPE versus training epoch.

In addition to training behavior, we further evaluate the forecasting performance of PatchSTG on the test set across different prediction steps.

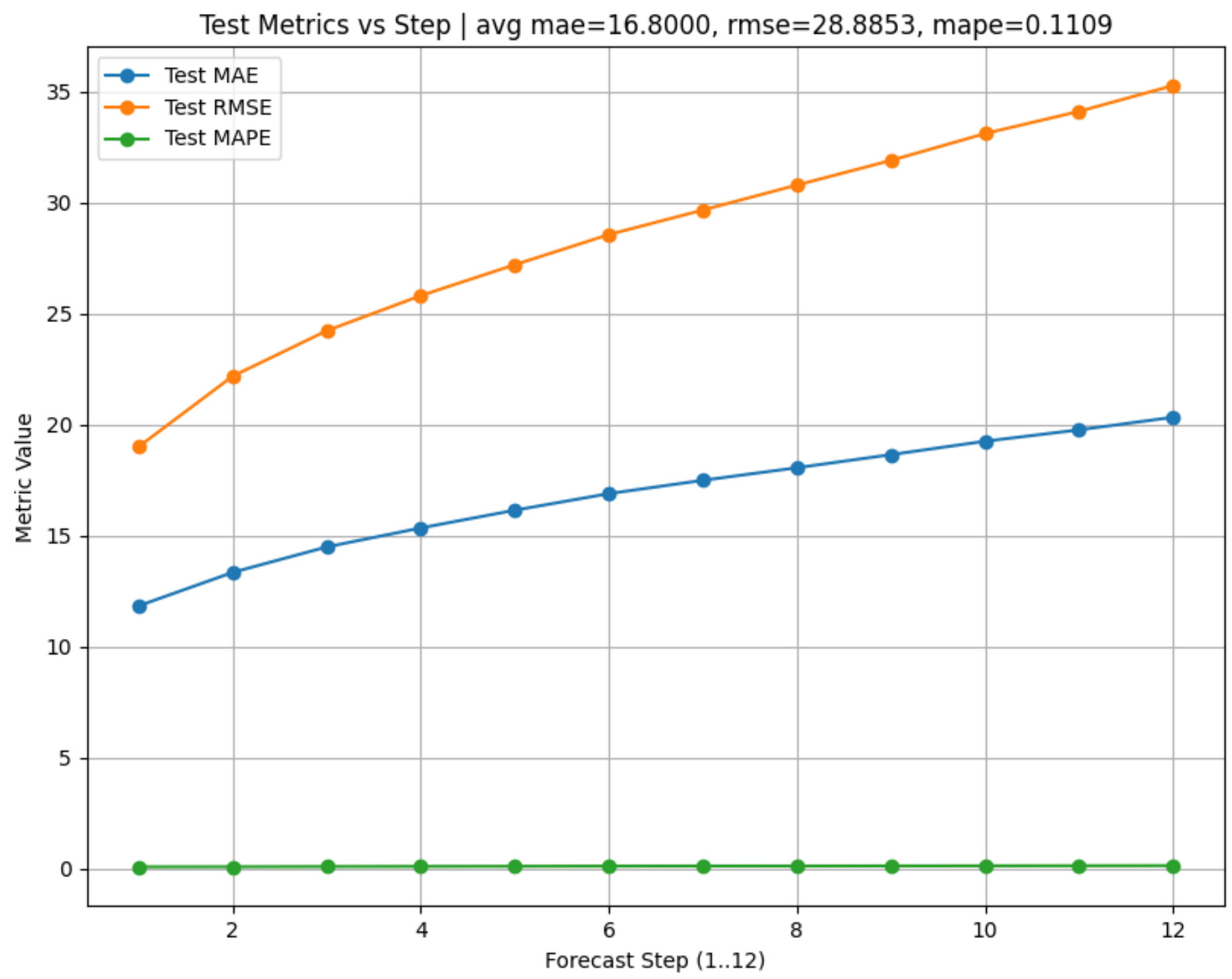

Figure 12: Test MAE, RMSE, and MAPE versus forecasting step. The proposed PatchSTG model maintains stable performance across multiple prediction horizons, with an average MAE of 16.80, RMSE of 28.89, and MAPE of 11.09%.

Figure 12 illustrates the test metrics (MAE, RMSE, and MAPE) with respect to the forecasting horizon. As the prediction step increases, the error gradually accumulates, which is consistent with common observations in multi-step traffic forecasting tasks. Nevertheless, the model maintains stable performance across all prediction steps. On average, PatchSTG achieves an MAE of 16.80, an RMSE of 28.89, and a MAPE of 11.09% on the test set, demonstrating its effectiveness in both short-term and longer-horizon traffic prediction.

## 5.3 Ablation Study

To assess the contribution of different components in PatchSTG, we conduct an ablation study on multiple large-scale traffic datasets. Table 2 summarizes the average forecast performance under various model configurations.

Table 2: Ablation study of PatchSTG on average results of large-scale traffic datasets. Bold indicates the best performance.

| Model Variant | SD MAE | SD RMSE | SD MAPE (%) | GBA MAE | GBA RMSE | GBA MAPE (%) | GLA MAE | GLA RMSE | GLA MAPE (%) | CA MAE | CA RMSE | CA MAPE (%) |
|---|---|---|---|---|---|---|---|---|---|---|---|---|
| w/o FGGC | 16.37 | 26.44 | 10.46 | 18.72 | 32.28 | 14.51 | 17.49 | 31.95 | 11.69 | 14.63 | 29.04 | 11.05 |
| w/o Depth | 16.18 | 26.91 | 10.28 | 18.63 | 32.33 | 14.11 | 17.38 | 32.05 | 11.73 | 14.68 | 29.32 | 11.09 |
| w/o Breadth | 16.69 | 27.55 | 10.27 | 18.90 | 32.44 | 14.28 | 17.65 | 32.66 | 12.18 | 15.04 | 29.74 | 11.17 |
| w/ PadDis | 16.03 | 26.42 | 10.34 | 18.60 | 32.32 | 14.03 | 17.17 | 31.80 | 10.63 | 14.50 | 29.14 | 11.00 |
| w/o PadSim | 16.05 | 26.43 | 10.25 | 18.57 | 32.34 | 14.32 | 18.57 | 32.18 | 11.80 | 14.87 | 29.52 | 11.11 |
| w/ METIS | 16.04 | 27.69 | 10.26 | 18.91 | 32.38 | 14.16 | 17.63 | 32.57 | 11.86 | 15.02 | 39.53 | 11.02 |
| w/ KMeans | 17.03 | 27.76 | 10.87 | 19.16 | 32.93 | 14.87 | 18.85 | 33.07 | 11.90 | 15.26 | 29.08 | 11.05 |
| w/o LKDT | 16.47 | 26.38 | 10.45 | 19.14 | 33.18 | 14.91 | 18.88 | 33.47 | 10.31 | 15.27 | 29.97 | 11.00 |
| **PatchSTG (Full)** | **15.53** | **26.04** | **9.26** | **18.32** | **31.23** | **13.24** | **17.93** | **30.23** | **10.30** | **14.65** | **28.60** | **11.65** |

Removing key components such as feature-guided graph construction, depth or breadth modeling leads to consistent performance degradation across all datasets and evaluation metrics. In addition, replacing the proposed partitioning strategy with alternative methods such as METIS or KMeans also results in inferior performance. These results demonstrate that each component of PatchSTG contributes positively to the overall forecasting accuracy, and the full model achieves the best performance.

### 5.4 Summary of Experimental Results

Overall, the experimental results demonstrate that the traffic dataset exhibits strong temporal periodicity and irregular spatial structure. The proposed PatchSTG model can be trained stably and effectively leverages spatial partitioning and attention mechanisms. Ablation further validates the necessity of each model component, confirming the effectiveness of the proposed design.

## 6. Discussion

The experimental results provide several important insights into the characteristics of the traffic data and the behavior of the proposed PatchSTG model.

First, the exploratory data analysis highlights strong temporal regularity and spatial heterogeneity in the Rhode Island traffic dataset. Daily and weekly traffic patterns exhibit clear periodic trends, while traffic intensity and variability differ substantially across sensors and regions. These observations confirm that traffic forecasting is inherently a spatiotemporal problem and motivate the need for models that can jointly capture temporal dynamics and spatial dependencies.

Second, the training and test results demonstrate that PatchSTG can be trained stably and maintains consistent performance across multiple forecasting horizons. The gradual increase in error with respect to prediction steps is expected in multi-step forecasting tasks, yet the model shows no abrupt performance degradation. This indicates that the learned representations are robust and generalize reasonably well to longer-term predictions.

Third, the ablation study provides empirical evidence for the effectiveness of each major component in PatchSTG. Removing spatial partitioning or local and global attention mechanisms consistently leads to worse performance across all regional subsets and evaluation metrics. In particular, the full PatchSTG model achieves the best results, suggesting that combining irregular spatial partitioning with dual attention is beneficial for capturing both local and long-range traffic interactions.

Despite these positive results, several limitations remain. The experiments focus on traffic data from a single state, and the regional subsets are derived from the same underlying sensor network. While this setup is sufficient for a course project, future work could evaluate the model on datasets from different cities or countries to further assess generalization. In addition, the current study emphasizes predictive accuracy, whereas other practical considerations such as real-time deployment efficiency or interpretability for traffic operators are not explored in depth.

Overall, these findings suggest that PatchSTG provides an effective approach for traffic forecasting under irregular spatial settings, while also highlighting several promising directions for future improvement.

## 7. Conclusion

In this work, we investigate the problem of traffic forecasting using real-world traffic sensor data from Rhode Island. Through exploratory data analysis, we demonstrate that the dataset exhibits strong temporal periodicity and highly irregular spatial structure, both of which pose challenges for conventional modeling approaches.

To address these challenges, we implemented and evaluated PatchSTG, a patch-based spatiotemporal forecasting model that combines irregular spatial partitioning with a dual attention encoder. Experimental results show that the model can be trained stably, achieves consistent performance across multiple forecasting horizons, and benefits from both local and global spatial modeling. Ablation studies further confirm that each major component of the model contributes positively to overall performance.

While this study is conducted as a student project and does not aim to achieve state-of-the-art performance, the results demonstrate the potential of patch-based spatiotemporal modeling for traffic forecasting. Future work could extend this project by incorporating additional datasets from different regions, improving model efficiency for real-time applications, or enhancing interpretability to better support traffic management decisions.

In summary, this work presents a complete pipeline from data analysis to model design and evaluation, and provides useful insights into the application of deep learning methods for real-world traffic forecasting problems.